\documentclass[review]{elsarticle}

\journal{Journal of \LaTeX\ Templates}
\usepackage{xcolor}
\usepackage{amsmath}
\renewcommand{\vec}[1]{\mathbf{#1}}
\usepackage[ruled]{algorithm2e}
\usepackage{multirow} 
\usepackage{graphicx}
\usepackage{subfig}
\usepackage{threeparttable}
\usepackage{caption}
\usepackage{algorithmic}
\renewcommand{\vec}[1]{\mathbf{#1}}

\usepackage[utf8x]{inputenc}
\DeclareUnicodeCharacter{ +2113}{~}
\begin{document}

\begin{frontmatter}

\title{Low-rank Kernel Learning for Graph-based Clustering}

\author{Zhao Kang, Liangjian Wen, Wenyu Chen,  Zenglin Xu}

\address{ School of Computer Science and Engineering\\
 University of Electronic Science and Technology of China, Chengdu, Sichuan, 611731 China}
%
%

%

\begin{abstract}
Constructing the adjacency graph is fundamental to graph-based clustering. Graph learning in kernel space has shown impressive performance on a number of benchmark data sets. However, its performance is largely determined by the choosed kernel matrix. To address this issue, previous multiple kernel learning algorithm has been applied to learn an optimal kernel from a group of predefined kernels. This approach might be sensitive to noise and limits the representation ability of the consensus kernel. In contrast to existing methods, we propose to learn a low-rank kernel matrix which exploits the similarity nature of the kernel matrix and seeks an optimal kernel from the neighborhood of candidate kernels. By formulating graph construction and kernel learning in a unified framework, the graph and consensus kernel can be iteratively enhanced by each other. Extensive experimental results validate the efficacy of the proposed method.   

\end{abstract}


\end{frontmatter}

\titlepage

\section{Introduction}
Clustering is a fundamental and important technique in machine learning, data mining, and pattern recognition \cite{jain1999data,zhu2018yotube,huang2018robust}. It aims to divide data samples into certain clusters such that similar objects lie in the same group. It has been utilized in various domains, such as image segmentation \cite{felzenszwalb2004efficient}, gene expression analysis \cite{jiang2004cluster}, motion segmentation \cite{elhamifar2009sparse}, image clustering \cite{yang2015class}, heterogeneous data analysis \cite{liu2015spectral}, document clustering \cite{yan2017novel,huang2018adaptive}, social media analysis \cite{he2014comment}, subspace learning \cite{gao2015multi,chen2012fgkm}. During the past decades, clustering has been extensively studied and many clustering methods have been developed, such as K-means clustering \cite{macqueen1967some,chen2013twkm}, spectral clustering \cite{von2007tutorial,kang2018unified}, subspace clustering \cite{liu2010robust,peng2018structured}, hierarchical clustering \cite{johnson1967hierarchical}, matrix factorization-based algorithms \cite{ding2010convex,huang2018self,huang2018robustDMKD}, graph-based clustering \cite{huang2015new,xuan2015topic}, and kernel-based clustering \cite{kang2017twin}. Among them, the K-means and spectral clustering are especially popular and have been extensively applied in practice. 

Basically, the K-means method iteratively assigns data points to their closest clusters and updates cluster centers. Nonetheless, it can not partition arbitrarily shaped clusters and is notorious for its sensitivity to the initialization of cluster centers \cite{ng2002spectral}. Later, the kernel K-means (KKM) was proposed to characterize data nonlinear structure information \cite{scholkopf1998nonlinear}. However, the user has to specify a kernel matrix as input, i.e., the user must assume a certain shape of the data distribution which is generally unknown. Consequently, the performance of KKM is highly dependent on the choice of the kernel matrix. This will be a stumbling block for the practical use of kernel method in real applications. This issue is partially alleviated by multiple kernel learning (MKL) technique which lets an algorithm do the picking or combination from a set of candidate kernels \cite{liu2017optimal,zhou2015recovery}. Since the kernels might be corrupted due to the contamination of the original data with noise and outliers. Thus, the induced kernel might still not be optimal \cite{gonen2011multiple}. Moreover, enforcing the optimal kernel being a linear combination of base kernels could lead to limited representation ability of the optimal kernel. Sometimes, MKL approach indeed performs worse than a single kernel method \cite{kang2018self}.

Spectral clustering, another classic method, presents more capability in detecting complex structures of data compared to other clustering methods \cite{yang2015multitask,yang2017discrete}. It works by embedding the data points into a vector space that is spanned by the spectrum of affinity matrix (or data similarity matrix). Therefore, the quality of the similarity graph is crucial to the performance of spectral clustering algorithm. Previously, the Gaussian kernel function is usually employed to build the graph matrix. Unfortunately, how to select a proper Gaussian parameter is an open problem \cite{zelnik2004self}. Moreover, the Gaussian kernel function is sensitive to noise and outliers. 

Recently, some advanced techniques have been developed to construct better similarity graphs. For instance, Zhu et al. \cite{zhu2014constructing} used a random forest-based method to identify discriminative features, so that subtle and weak data affinity can be captured. More importantly, adaptive neighbors method \cite{nie2014clustering} and self-expression approach \cite{kang2017kernel} have been proposed to learn a graph automatically from the data. This automatic strategy can tackle data with structures at different scales of size and density and often provides a high-quality graph, as demonstrated in clustering \cite{nie2014clustering,patel2014kernel}, semi-supervised classification \cite{zhuang2012non,li2015learning}, and many others.

In this paper, we learn the graph in kernel space. To address the kernel dependence issue, we develop a novel method to learn the consensus kernel. Finally, a unified model which seamlessly integrates graph learning and kernel learning is proposed. On one hand, the quality of the graph will be enhanced if it is learned with an adaptive kernel. On the other hand, the learned graph will help to improve the kernel learning since graph and kernel are the same in essence in terms of the pairwise similarity measure. 

The main novelty of this paper is revealing the underlying structure of the kernel matrix by imposing a low-rank regularizer on it. Moreover, we find an ideal kernel in the neighborhood of base kernels, which can improve the robustness of the learned kernel. This is beneficial in practice since the candidate kernels are often corrupted. Consequently, the optimal kernel can reside in some kernels' neighborhood. In summary, we highlight the main contributions of this paper as follows: 
\begin{itemize}
\item We propose a unified model for learning an optimal consensus kernel and a similarity graph matrix, where the result of one task is used to improve the other one. In other words, we consider the possibility that these two learning processes may need to negotiate with each other to achieve the overall optimality. 
\item By assuming the low-rank structure of the kernel matrix, our model is in a better position to deal with real data. Instead of enforcing the optimal kernel being a linear combination of predefined kernels, our model allows the most suitable kernel to reside in its neighborhood.
\item Extensive experiments are conducted to compare the performance of our proposed method with existing state-of-the-art clustering methods. Experimental results demonstrate the superiority of our method.
\end{itemize}

The rest of the paper is organized as follows. Section \ref{related} describes related works. Section \ref{proposed} introduces the proposed graph and kernel learning method. Experimental results and analysis are presented in Section \ref{experiment}. Section \ref{conclusion} draws conclusions.

\textbf{Notations.} Given a data matrix $X\in\mathcal{R}^{m \times n}$ with $m$ features and $n$ samples, we denote its $(i,j)$-th element and $i$-th column as $x_{ij}$ and $x_i$, respectively. The $\ell_2$-norm of vector \textbf{$x$} is represented by \textbf{$\|x\|=\sqrt{x^T\cdot x}$}, where $\textbf{$x^T$}$ is the transpose of \textbf{$x$}. The $\ell_1$-norm of $X$ is denoted by $\|X\|_1=\sum_{ij}|x_{ij}|$.  The squared Frobenius norm is defined as $\|X\|_F^2=\sum_{ij}x_{ij}^2$. The definition of $X$'s nuclear norm is $\|X\|_*=\sum_i\sigma_i$, where $\sigma_i$ is the $i$-th singular value of $X$. $I$ represents the identity matrix with proper size. $Tr(\cdot)$ denotes the trace operator. $Z\geq 0$ means all elements of $Z$ are nonnegative. Inner product is denoted by $<x_i, x_j>=x_i^T\cdot x_j$.

\section{Related Work}
\label{related}
To cope with noise and outliers, robust kernel K-means (RKKM) \cite{du2015robust} algorithm has been proposed recently. In this method, the squared $\ell_2$-norm of error construction term is replaced by the $\ell_{2,1}$-norm. RKKM demonstrates compelling performance on a number of benchmark data sets. To alleviate the efforts for exhaustive search of the most suitable kernel on a pre-specified pool of kernels, the authors further proposed a robust multiple kernel K-means (RMKKM) algorithm. RMKKM conducts robust K-means by learning an appropriate consensus kernel from a linear combination of multiple candidate kernels. It shows that RMKKM has great potential to integrate complementary information from different sources along with heterogeneous features \cite{yu2012optimized}. This leads to better performance of RMKKM than that of RKKM.

As aforementioned, the graph-based clustering methods have achieved impressive performance. To resolve the graph construction challenge, simplex sparse representation (SSR) \cite{huang2015new} was proposed to learn the affinity between pairs of samples. It is based on the so-called self-expression property, i.e., each data point can be represented as a weighted combination of other points \cite{liu2010robust}. More similar data points will receive larger weights. Therefore, the induced weight matrix reveals the relationships between data points and encodes the data structure. Next, the learned affinity graph matrix is inputted to the spectral clustering algorithm. Empirical experiments demonstrate the superior performance of this approach.

Recently, Kang et al. \cite{kang2017twin} have proposed to learn the similarity matrix in kernel space based on self-expression. They built a joint framework for similarity matrix construction and cluster label learning. Both single kernel method (SCSK) and multiple kernel approach (SCMK) were developed. They learn an optimal kernel using the same way as adopted by RMKKM. In specific, SCMK and RMKKM directly replace the kernel matrix in single kernel model with a combined kernel, which is expressed as a linear combination of pre-specified kernels in the constraint. This is a straightforward way and also a popular approach in the literature. However, it ignores the structure information of the kernel matrix. In essence, the kernel matrix is a measure of pairwise similarity between data points. Hence, the kernel matrix is low-rank in general \cite{xia2014robust}. Moreover, they strictly require that the optimal kernel is a linear combination of base kernels. This might limit its realistic application since real-world data is often corrupted and the ideal kernel might reside in the neighborhood of the combined kernel. Besides, it is time-consuming and impractical to design a large pool of kernels. Hence it is impossible to obtain a globally optimal kernel. What we can do is to find a way to make the best use of candidate kernels. 

In this paper, we propose to learn a similarity graph and kernel matrix jointly by exploring the kernel matrix structure. With the low-rank requirement on the kernel matrix, we are expected to exploit the similarity nature of the kernel matrix. Different from existing methods, we relax the strict condition that the optimal kernel is a linear combination of predefined kernels in order to account noise in real data. This enlarges the region from which an ideal kernel can be chosen and therefore is in a better position than the previous approach to finding a more suitable kernel. In particular, in a similar spirit of  robust principal component analysis (RPCA) \cite{kang2015robust}, the combined kernel is factorized into a low-rank component (optimal kernel matrix) and residual.

\section{Proposed Methodology}
\label{proposed}
\subsection{Formulation}
In general, the self-expression based graph learning problem can be formulated as
\begin{equation}
\min_{Z} \frac{1}{2}\|X-XZ\|_F^2+\alpha \rho(Z)\quad  s.t.\quad  Z\geq 0,
\label{original}
\end{equation}
where $\alpha>0$ is a regularization parameter, self-expression coefficient $Z$ is often assumed to be nonnegative, $\rho(Z)$ is the regularization term on $Z$. Two commonly used assumptions about $\rho(Z)$ are low-rank and sparse, corresponding to $\|Z\|_*$ and $\|Z\|_1$ respectively. 
Suppose $\phi: \mathcal{R}^D\rightarrow\mathcal{H}$ maps the data points from the input space to a reproducing kernel Hilbert space $\mathcal{H}$. Then, based on the kernel trick, the $(i,j)$-th element of kernel matrix $H$ is $H_{ij}=<\phi(x_i),\phi(x_j)>$. In kernel space, Eq. (\ref{original}) gives
\begin{equation}
\begin{split}
&\min_{Z} \frac{1}{2}\|\phi(X)-\phi(X)Z\|_F^2+\alpha  \rho(Z)\Longleftrightarrow\\
\min_{Z} & \frac{1}{2}\!Tr(\!\phi(X)^T\!\phi(X)\!-\!\phi(X)^T\!\phi(X)\!Z\!-\!Z^T\!\phi(X)^T\!\phi(X)\!+\!Z^T\!\phi(X)^T\!\phi(X) Z\!)\!+\!\alpha\! \rho(Z),\\
\Longleftrightarrow
& \min_{Z} \frac{1}{2}Tr(H-2HZ+Z^T HZ)+\alpha  \rho(Z)\quad s.t. \quad Z\ge0.
\label{kernelsparse}
\end{split}
\end{equation}
This model is capable of recovering the linear relationships among the data samples in the new space, and thus the nonlinear relationships in the original representation. One limitation of Eq. (\ref{kernelsparse}) is that its performance will heavily depend on the inputted kernel matrix. To overcome this drawback, we can learn a suitable kernel $K$ from $r$ predefined kernels $\{H^i\}_{i=1}^{r}$. Different from existing MKL method, we aim to increase the consensus kernel's representation ability by considering noise effect. Finally, our proposed \textbf{L}ow-rank \textbf{K}ernel learning for \textbf{G}raph matrix (LKG) is formulated as following
\begin{equation}
\begin{split}
\min_{Z, K, \vec{g}}\hspace{.1cm} & \frac{1}{2}Tr(\!K\!-\!2KZ\!+\!Z^TKZ\!)\!+\!\alpha \rho(Z)\!+\!\beta\! \|K\|_*\!+\!\gamma\|\!K\!-\!\sum\limits_i g_iH^i\|_F^2 \\
& s.t.\quad Z\geq 0, \quad K\geq 0,\quad g_i\geq 0, \quad \sum\limits_{i=1}^r {g_i}=1,
\label{LKG}
\end{split}
\end{equation}
where $g_i$ is the weight for kernel $H^i$, kernel matrix $K$ is nonnegative, the constraints for $g$ are from standard MKL method. If a kernel is not appropriate due to the bad choice of metric or parameter, or a kernel is severely corrupted by noise or outliers, the corresponding $g_i$ will be assigned a small value.

In Eq. (\ref{LKG}), $\|K\|_*$ explores the structure of the kernel matrix, so that the learned $K$ will respect the correlations among samples, i.e., the cluster structure of data. Moreover, enforcing the nuclear norm regularizer on $K$ will make $K$ robust to noise and errors. The last term in Eq. (\ref{LKG}) means that we seek an optimal kernel $K$ in the neighborhood of $\sum\limits_i g_iH^i$, which makes our model in a better position than the previous approach to identify a more suitable kernel. Due to noise and outliers, $\sum\limits_i g_iH^i$ could be a noisy observation of the ideal kernel $K$. As a matter of fact, this is similar to RPCA \cite{candes2011robust,kang2015robustpca}, where the original noise data is decomposed into a low-rank part and an error part. Formulating $Z$ and $K$ learning in a unified model reinforces the underlying connections between learning the optimal kernel and graph learning. By iteratively updating $Z$, $K$, $\vec{g}$, they can be repeatedly improved.

\subsection{Optimization }
We propose to solve the problem (\ref{LKG}) based on the alternating direction method of multipliers (ADMM) \cite{boyd2011distributed}. First, we introduce two auxiliary variables to make variables separable and rewrite the problem (\ref{LKG}) in the following equivalent form 
\begin{equation}
\begin{split}
 &\min_{Z, K, \vec{g}}\hspace{.1cm} \frac{1}{2}Tr(\!K\!-\!2KZ\!+\!Z^TKZ\!)\!+\!\alpha \rho(J)\!+\!\beta\! \|W\|_*\!+\!\gamma\|\!K\!-\!\sum\limits_i g_iH^i\|_F^2 \\
& s.t.\quad Z\geq 0, \quad K\geq 0,\quad g_i\geq 0, \quad \sum\limits_{i=1}^r {g_i}=1,\quad J=Z,\quad W=K.
\label{newLKG}
\end{split}
\end{equation}
The corresponding augmented Lagrangian function is
\begin{equation}
\begin{split}
&\mathcal{L}(Z, K, \vec{g}, J, W, Y_1, Y_2)=\frac{1}{2}Tr(\!K\!-\!2KZ\!+\!Z^TKZ\!)\!+\!\alpha \rho(J)\!+\!\beta\! \|W\|_*+\\
&\gamma\|K-\sum\limits_i g_iH^i\|_F^2 +\frac{\mu}{2}\left(\|J-Z+\frac{Y_1}{\mu}\|_F^2+\|W-K+\frac{Y_2}{\mu}\|_F^2\right),
\label{lag}
\end{split}
\end{equation}
where $\mu >0$ is a penalty parameter and $ Y_1$, $Y_2$ are lagrangian multipliers. These variables can be updated alternatingly, one at each step, while keeping the others fixed.

To solve Z, the objective function (\ref{lag}) becomes
\begin{equation}
\min_Z \frac{1}{2}Tr(\!K\!-\!2KZ\!+\!Z^TKZ\!)\!+\frac{\mu}{2}\|J-Z+\frac{Y_1}{\mu}\|_F^2.
\end{equation}
It can be solved by setting its first derivative to zero. Then we have
\begin{equation}
 Z\!=\!(\!K\!+\!\mu I\!)^{-1}(\!K\!+\!\mu J\!+\!Y_1\!).
\label{updateZ}
\end{equation}
Similarly, we can obtain the updating rule for $K$ as
\begin{equation}
K\!=\frac{\!2\gamma \sum\limits_i g_iH^i\!+\!\mu W\!+\!Y_2\!-\!\frac{I}{2}\!+\!Z^T\!-\!\frac{ZZ^T}{2}\!}{\mu+2\gamma}.
\label{updateK}
\end{equation}

To solve $J$, the sub-problem is 
\begin{equation}
\min_J  \alpha \rho(J)\!+\frac{\mu}{2}\|J-Z+\frac{Y_1}{\mu}\|_F^2.
\end{equation}
Depending on the regularization strategy, we obtain different closed-form solutions for $J$. Let's define $D=Z-\frac{Y_1}{\mu}$ and write the singular value decomposition (SVD) of $D$ as $Udiag(\sigma)V^T$. Then, for low-rank representation, it yields 
 \begin{equation}
 J=U diag(\textit{max}\{\sigma-\frac{\alpha}{\mu},0\}) V^T.
\label{lowrank}
\end{equation}
To obtain a sparse representation, we can update $J$ elemently as 
\begin{equation}
 J_{ij}=max(|D_{ij}|-\frac{\alpha}{\mu},0)\cdot \textit{sign}(D_{ij}).
\label{sparse}
\end{equation}

To solve $W$, we have
\begin{equation}
\min_W  \beta \|W\|_*+\frac{\mu}{2}\|W-K+\frac{Y_2}{\mu}\|_F^2.
\end{equation}
By letting $G=K-\frac{Y_2}{\mu}$ and $\textit{SVD}(G)=\bar{U}diag(\bar{\sigma})\bar{V}^T$, then we have
 \begin{equation}
W=\bar{U} diag(max\{\bar{\sigma}-\frac{\beta}{\mu},0\}) \bar{V}^T.
\label{lowrankW}
\end{equation}

To solve $\vec{g}$, the optimization problem (\ref{LKG}) becomes
\begin{equation}
\min_\vec{g}\gamma\vec{g}^TM\vec{g}\!-\!\vec{a}^T\vec{g}\quad s.t.\quad g_i\geq 0, \quad \sum\limits_{i=1}^r {g_i}=1,
\label{solveg}
\end{equation}
where $M_{ij}=\!Tr\!(\!H^i H^j\!)$ and $a_i\!=\!\frac{\gamma}{2}Tr\!(\!KH^i\!)$. It is a Quadratic Programming problem with linear constraints, which can be easily solved with existing packages. In sum, our algorithm for solving the problem (\ref{LKG}) is outlined in Algorithm 1. 

After obtaining the graph $Z$, we can use it to do clustering, semi-supervised classification, and so on. In this work, we focus on the clustering task. Specifically, we run the spectral clustering \cite{ng2002spectral} algorithm on $Z$ to achieve the final results.
\begin{algorithm}
\caption{The algorithm to solve  (\ref{LKG})}
\label{alg2}
 {\bfseries Input:} Kernel matrices $\{H^i\}_{i=1}^{i=r}$, parameters $\alpha>0$, $\beta>0$, $\gamma>0$, $\mu>0$.\\
{\bfseries Initialize:} Random matrix $J$, $Y_1=Y_2=Y_3=0$, $g_i=1/r$, $K=W=\sum\limits_{i=1}^r g_iH^i/r$.\\
 {\bfseries REPEAT}
\begin{algorithmic}[1]
\STATE Calculate $Z$ by (\ref{updateZ}).
\STATE $Z$=max($Z$, 0).
 \STATE Update $K$ according to (\ref{updateK}).
\STATE $K$=max($K$, 0).
\STATE Calculate $J$ using (\ref{sparse}) or (\ref{lowrank}).
\STATE $J$=max($J$, 0).
\STATE Calculate $W$ using (\ref{lowrankW}).
\STATE $W$=max($W$, 0).
\STATE Solve $\vec{g}$ using (\ref{solveg}).
\STATE Update Lagrange multipliers $Y_1$ and $Y_2$ as
\begin{eqnarray*}
Y_1=Y_1+\mu(J-Z),\\
Y_2=Y_2+\mu(W-K).
\end{eqnarray*}
\end{algorithmic}
\textbf{ UNTIL} {stopping criterion is met.}
\end{algorithm}
\subsection{Complexity Analysis}
 The time complexity for each kernel construction is $\mathcal{O}(n^2)$. The computational cost for $Z$ and $K$ is $\mathcal{O}(n^3)$. For $W$, it requires an SVD for every iteration and its complexity is $\mathcal{O}(n^3)$, which can be $\mathcal{O}(kn^2)$ if we employ partial SVD ($k$ is the lowest rank we can find) based on package PROPACK \cite{larsen2004propack}. For $J$, depending on the choice of regularizer, we have different complexity. For low-rank representation, it is the same as $W$. The complexity of obtaining a sparse solution $J$ is $\mathcal{O}(n^2)$. It is a quadratic programing problem for $\vec{g}$, which can be solved in polynomial time. Fortunately, the size of $\vec{g}$ is a small number $r$. The updating of $Y_1$ and $Y_2$ cost $\mathcal{O}(n^2)$.

\begin{figure*}
\centering
\subfloat[BA\label{ba}]{\includegraphics[width=.33\textwidth]{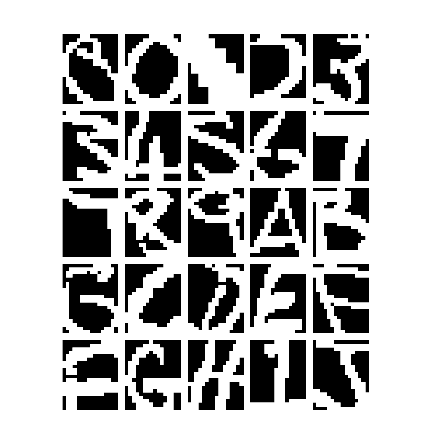}}
\subfloat[YALE\label{yale}]{\includegraphics[width=.33\textwidth]{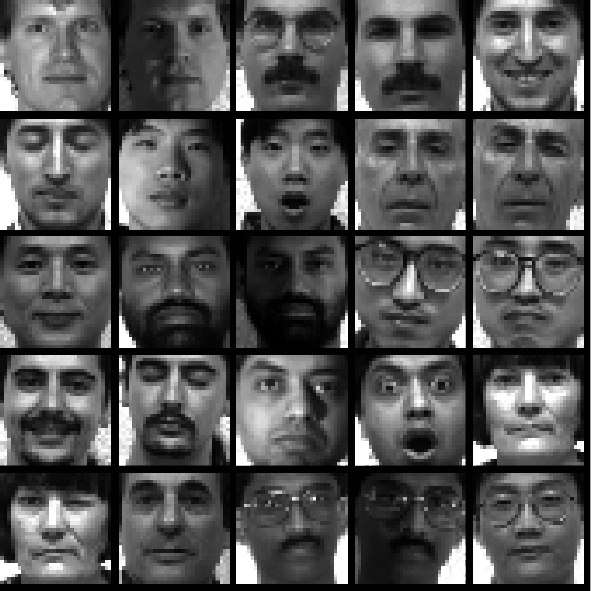}}
\subfloat[JAFFE\label{jaffe}]{\includegraphics[width=.33\textwidth]{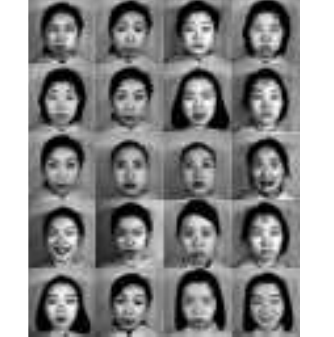}}
\caption{Sample images of BA, YALE and JAFFE.}
\end{figure*}
\section{Experiments}
\label{experiment}
\subsection{Data Sets}
We examine the effectiveness of our method using eight real-world benchmark data sets, which are commonly used in the literature. The basic information of data sets is shown in Table \ref{data}. In specific, the first five data sets are images, and the other four are text corpora\footnote{http://www-users.cs.umn.edu/~han/data/tmdata.tar.gz}\footnote{http://www.cad.zju.edu.cn/home/dengcai/Data/TextData.html}. 

Five image data sets include four famous face databases (ORL\footnote{http://www.cl.cam.ac.uk/research/dtg/attarchive/facedatabase.html}, YALE\footnote{http://vision.ucsd.edu/content/yale-face-database}, AR\footnote{http://www2.ece.ohio-state.edu/~aleix/ARdatabase.html} and JAFFE\footnote{http://www.kasrl.org/jaffe.html}), and a binary alpha digits data set BA\footnote{http://www.cs.nyu.edu/~roweis/data.html}. As shown in Figure \ref{ba}, BA consists of digits of ``0" through ``9" and letters of capital ``A" through ``Z". In YALE, ORL, AR, and JAFEE, each image has different facial expressions or configurations due to times, illumination conditions, and glasses/no glasses. Hence, these data sets are contaminated at different levels. Figure \ref{yale} and \ref{jaffe}  show some example images from YALE and JAFFE database. 

Following the setting in \cite{kang2017twin}, we manually construct 12 kernels. They consist of seven Gaussian kernels  $H(x,y)=exp(-\|x-y\|_2^2/(td_{max}^2))$ with $t\in\{0.01, 0.05, 0.1, 1, 10,$ $ 50, 100\}$, where $d_{max}$ denotes the maximal distance between data points; a linear kernel $H(x,y)=x^Ty$; four polynomial kernels $H(x,y)=(a+x^Ty)^b$ of the form with $a\in\{0,1\}$ and $b\in\{2,4\}$. In addition, all kernel matrices are normalized to $[0,1]$ range. This can be done through dividing each element by the largest element in its corresponding kernel matrix.
\captionsetup{position=top}
\begin{table}[!htbp]
\centering
\caption{Description of the data sets}
\label{data}
\begin{tabular}{|l|c|c|c|}
\hline
&\textrm{\# instances}&\textrm{\# features}&\textrm{\# classes}\\\hline
\textrm{YALE}&165&1024&15\\\hline
\textrm{JAFFE}&213&676&10\\\hline
\textrm{ORL}&400&1024&40\\\hline
\textrm{AR}&840&768&120\\\hline
\textrm{BA}&1404&320&36\\\hline
\textrm{TR11}&414&6429&9\\\hline
\textrm{TR41}&878&7454&10\\\hline
\textrm{TR45}&690&8261&10\\\hline
\textrm{TDT2}&9394&36771&30\\\hline
\end{tabular}
\end{table}

\subsection{Evaluation Metrics}

To quantitatively assess our algorithm's performance on the clustering task, we use the popular measures, i.e., accuracy (Acc) and normalized mutual information (NMI).

Acc discovers the one-to-one relationship between clusters and classes. Let $l_i$ and $\hat{l}_i$ be the clustering result and the ground truth cluster label of $x_i$, respectively. Then the Acc is defined by
\[
Acc=\frac{\sum_{i=1}^n \delta(\hat{l}_i, map(l_i))}{n},
\]
where $n$ is the total number of samples, delta function $\delta(x,y)$ equals one if and only if $x=y$ and zero otherwise, and map($\cdot$) is the best permutation mapping function that maps each cluster index to a true class label  based on Kuhn-Munkres algorithm.

The NMI measures the quality of clustering. Given two sets of clusters $L$ and $\hat{L}$,
\[
\textrm{NMI}(L,\hat{L})=\frac{\sum\limits_{l\in L,\hat{l}\in\hat{L}} p(l,\hat{l})\textrm{log}(\frac{p(l,\hat{l})}{p(l)p(\hat{l})})}{\textrm{max}(H(L),H(\hat{L}))},
\]  
where $p(l)$ and $p(\hat{l})$ represent the marginal probability distribution functions of $L$ and $\hat{L}$, respectively, induced from the joint distribution $p(l,\hat{l})$ of $L$ and $\hat{L}$. $H(\cdot)$ is the entropy function. The greater NMI means the better clustering performance.


\subsection{Comparison Methods}
To fully examine the effectiveness of our proposed algorithm, we compare with both graph-based clustering methods and kernel methods. More concretely, we have Kernel K-means (KKM) \cite{scholkopf1998nonlinear}, Spectral Clustering (SC) \cite{ng2002spectral}, Robust Kernel K-means (RKKM) \cite{du2015robust}, Simplex Sparse Representation (SSR) \cite{huang2015new} and SCSK \cite{kang2017twin}. Among them, SC, SSR, and SCSK are graph-based clustering methods. Since SSR is developed in the feature space, we only need run it once. For other techniques, we run them on each kernel and report their best performances as well as their average performances over those kernels. 

We also compare with a number of multiple kernel learning methods. We directly implement the downloaded programs of the  comparison methods on those 12 kernels: 

Multiple Kernel K-means (MKKM)\footnote{http://imp.iis.sinica.edu.tw/IVCLab/research/Sean/mkfc/code}. The MKKM \cite{huang2012multiple} is an extension of K-means to the situation when multiple kernels exist. 

Affinity Aggregation for Spectral Clustering (AASC)\footnote{http://imp.iis.sinica.edu.tw/IVCLab/research/Sean/aasc/code}. The AASC \cite{huang2012affinity} extends spectral clustering to deal with multiple affinities. 

Robust Multiple Kernel K-means (RMKKM)\footnote{https://github.com/csliangdu/RMKKM}. The RMKKM \cite{du2015robust} extends K-means to deal with noise and outliers in a multiple kernel setting. 

Twin learning for Similarity and Clustering with Multiple Kernel (SCMK) \cite{kang2017twin}. Recently proposed graph-based clustering method with multiple kernel learning capability. Both RMKKM and SCMK rigorously require that the consensus kernel is a combination of base kernels.

Low-rank Kernel learning for Graph matrix (LKG). Our proposed low-rank kernel learning for graph-based clustering method. After obtaining similarity graph matrix $Z$, we run the spectral clustering algorithm to finish the clustering task. We examine both low-rank and sparse regularizer and denote their corresponding methods as LKGr and LKGs, respectively.
 

\subsection{Results}
\begin{table*}[!ht]
\centering
\renewcommand{\arraystretch}{1.1}
\setlength{\tabcolsep}{.1pt}
\subfloat[Accuracy(\%)\label{acc}]{
\resizebox{1.0\textwidth}{!}{
\begin{tabular}{ |l  |c |c| c| c |c| | c| c| c| c |c| c| c| c| c|c|c }
\hline
Data 	& KKM		   & SC		      & RKKM	 &SSR     & SCSK			        & MKKM  & AASC  & RMKKM & SCMK 		      & LKGs		       & LKGr	   \\	\hline
YALE  	& 47.12(38.97) & 49.42(40.52) & 48.09(39.71)&54.55  & 55.85(45.35)           & 45.70 & 40.64 & 52.18 & 56.97            & 62.42 & \textbf{66.06} \\	\hline
JAFFE 	& 74.39(67.09) & 74.88(54.03) & 75.61(67.98)&87.32  & 99.83(86.64)  & 74.55 & 30.35 & 87.07 & \textbf{100.00}  & 98.12          & 98.60    \\	\hline
ORL   	& 53.53(45.93) & 57.96(46.65) & 54.96(46.88)&69.00  & 62.35(50.50)           & 47.51 & 27.20 & 55.60 & 65.25            & 71.5  & \textbf{73.50}	\\	\hline
AR   	& 33.02(30.89) & 28.83(22.22) & 33.43(31.20)&65.00  & 56.79(41.35)           & 28.61 & 33.23 & 34.37 & 62.38       	  & \textbf{65.83} & 60.47	\\	\hline
BA 		& 41.20(33.66) & 31.07(26.25) & 42.17(34.35) &23.97 & 47.72(39.50)           & 40.52 & 27.07 & 43.42 & 47.34       	  & 47.93 & \textbf{50.50}	\\	\hline
TR11   	& 51.91(44.65) & 50.98(43.32) & 53.03(45.04) &41.06 & 71.26(54.79)  & 50.13 & 47.15 & 57.71 & \textbf{73.43}   & 67.63          & 65.70       	\\	\hline
TR41  	& 55.64(46.34) & 63.52(44.80) & 56.76(46.80) &63.78 & \textbf{67.43}(53.13)  & 56.10 & 45.90 & 62.65 & 67.31	  & 62.64        & 63.44     	\\	\hline
TR45  	& 58.79(45.58) & 57.39(45.96) & 58.13(45.69) &71.45 & 74.02(53.38)           & 58.46 & 52.64 & 64.00 & 74.35      	  & 75.94 & \textbf{77.39}	\\	\hline
TDT2  	& 47.05(35.58) & 52.63(45.26) & 48.35(36.67) &20.86 & 55.74(44.82)             & 34.36 & 19.82 & 37.57 & 56.42     	  &58.77& \textbf{60.48}	\\	\hline
\end{tabular}
}

}\\
\renewcommand{\arraystretch}{1.1}
\subfloat[NMI(\%)\label{NMI}]{
\resizebox{1.0\textwidth}{!}{
\begin{tabular}{ |l  |c |c| c| c |c| | c| c| c| c |c| c| c| c| c|c|c  }
\hline
Data 	& KKM		    & SC	       	& RKKM	  &SSR       & SCSK			        & MKKM  & AASC  & RMKKM & SCMK 			& LKGs		        & LKGr	   \\	\hline
YALE 	& 51.34(42.07)  & 52.92(44.79)  & 52.29(42.87)&57.26  & 56.50(45.07)          & 50.06 & 46.83 & 55.58 & 56.52         & 61.72     & \textbf{64.57}	\\	\hline
JAFFE 	& 80.13(71.48)  & 82.08(59.35)  & 83.47(74.01)&92.93  & 99.35(84.67) & 79.79 & 27.22 & 89.37 & \textbf{100.00}&  97.00               & 98.73	\\	\hline
ORL 	& 73.43(63.36)  & 75.16(66.74)  & 74.23(63.91) &84.23 & 78.96(63.55)          & 68.86 & 43.77 & 74.83 & 80.04	        & 83.93   & \textbf{85.10}	\\	\hline
AR 		& 65.21(60.64)  & 58.37(56.05)  & 65.44(60.81) &84.16 & 76.02(59.70)        & 59.17 & 65.06 & 65.49 & 81.51	        & \textbf{84.69}    & 81.05	\\	\hline
BA  	& 57.25(46.49)  & 50.76(40.09)  & 57.82(46.91) &30.29 & 63.04(52.17) & 56.88 & 42.34 & 58.47 & 62.94	        & 60.12             & \textbf{63.20}	\\	\hline
TR11  	& 48.88(33.22)  & 43.11(31.39)  & 49.69(33.48) &27.60 & 58.60(37.58)          & 44.56 & 39.39 & 56.08 & 60.15       	& 62.30     & \textbf{63.50}	\\	\hline
TR41 	& 59.88(40.37)  & 61.33(36.60)  & 60.77(40.86)&59.56  & 65.50(43.18) & 57.75 & 43.05 & 63.47 & 65.11        	& \textbf{66.23}    & 61.78	\\	\hline
TR45	& 57.87(38.69)  & 48.03(33.22)  & 57.86(38.96)&67.82  & 74.24(44.36)          & 56.17 & 41.94 & 62.73 & 74.97& 70.97             & \textbf{75.22}	\\	\hline
TDT2  	& 55.28(38.47) & 52.23(27.16) & 54.46(42.19) &02.44 & 58.35(46.37)           & 41.36 & 02.14 & 47.13 &  59.84     	  & 60.75& \textbf{62.85}	\\	\hline
\end{tabular}
}
}\\
\caption{Performance of various clustering methods on benchmark data sets. For single kernel methods (The 1st, 2nd, 3rd, 5th columns), the average performance over those 12 kernels is put in parenthesis. The best results for these algorithms are highlighted in bold. \label{clusterres}}
\end{table*}
\captionsetup{position=bottom}
\begin{figure*}
\centering
\subfloat[$\alpha=10^{-5}$]{\includegraphics[width=.49\textwidth]{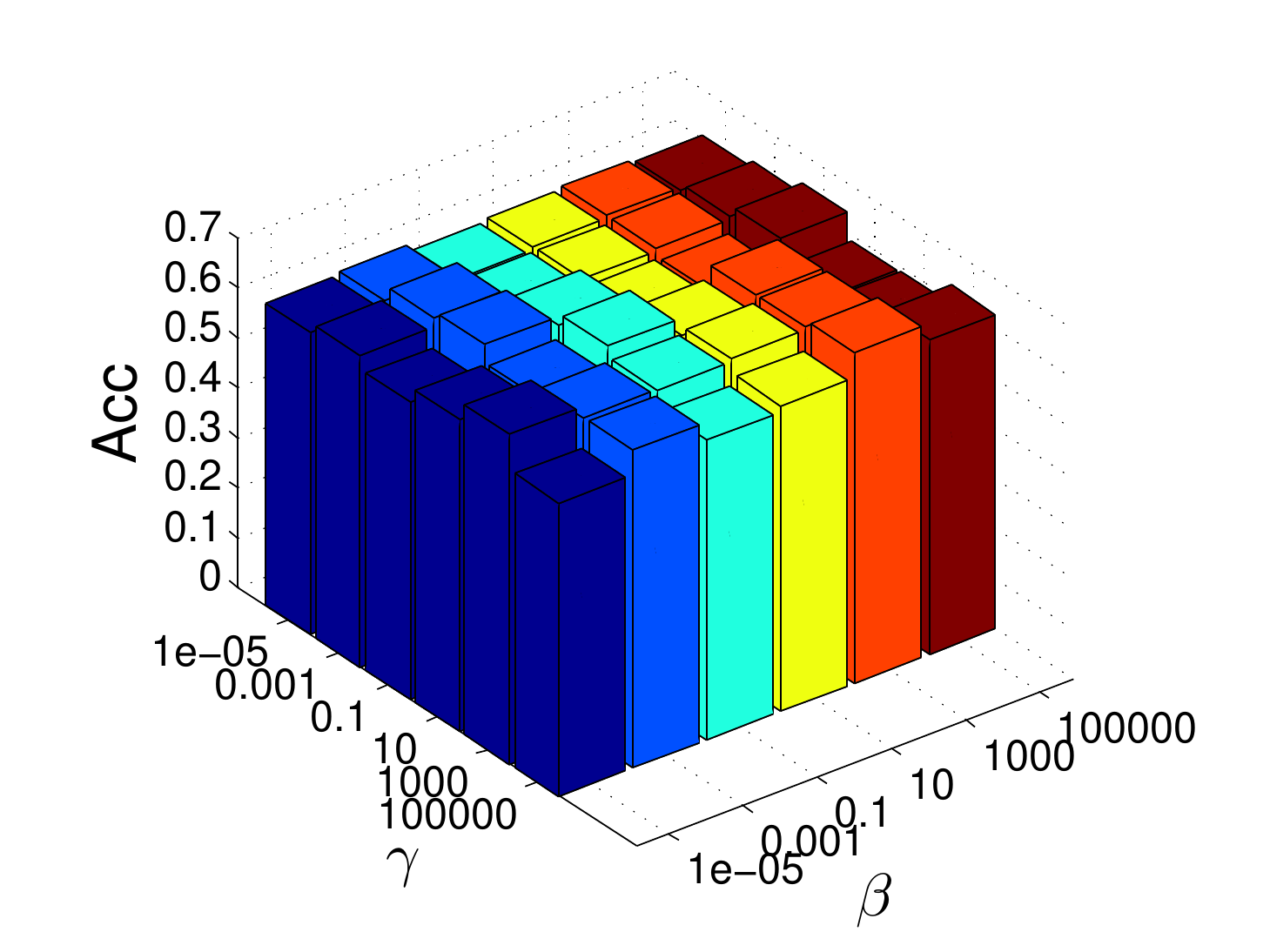}}
\subfloat[$\alpha=10^{-2}$]{\includegraphics[width=.49\textwidth]{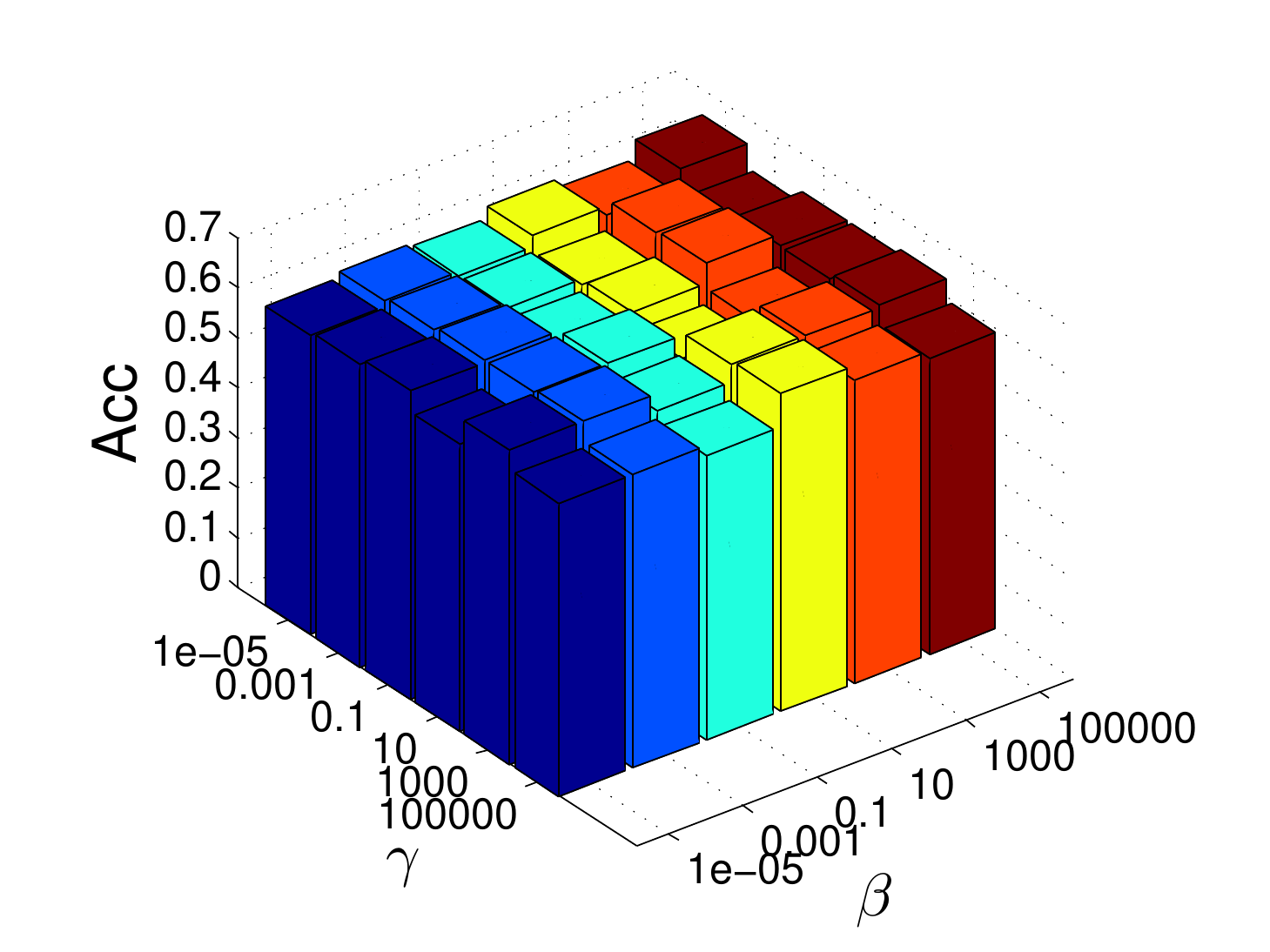}}
\caption{The clustering accuracy of LKGr on YALE Data w.r.t. $\gamma$ and $\beta$.}
\label{yalelkgs}
\end{figure*}

\begin{figure*}
\centering
\subfloat[$\alpha=10^{-5}$]{\includegraphics[width=.49\textwidth]{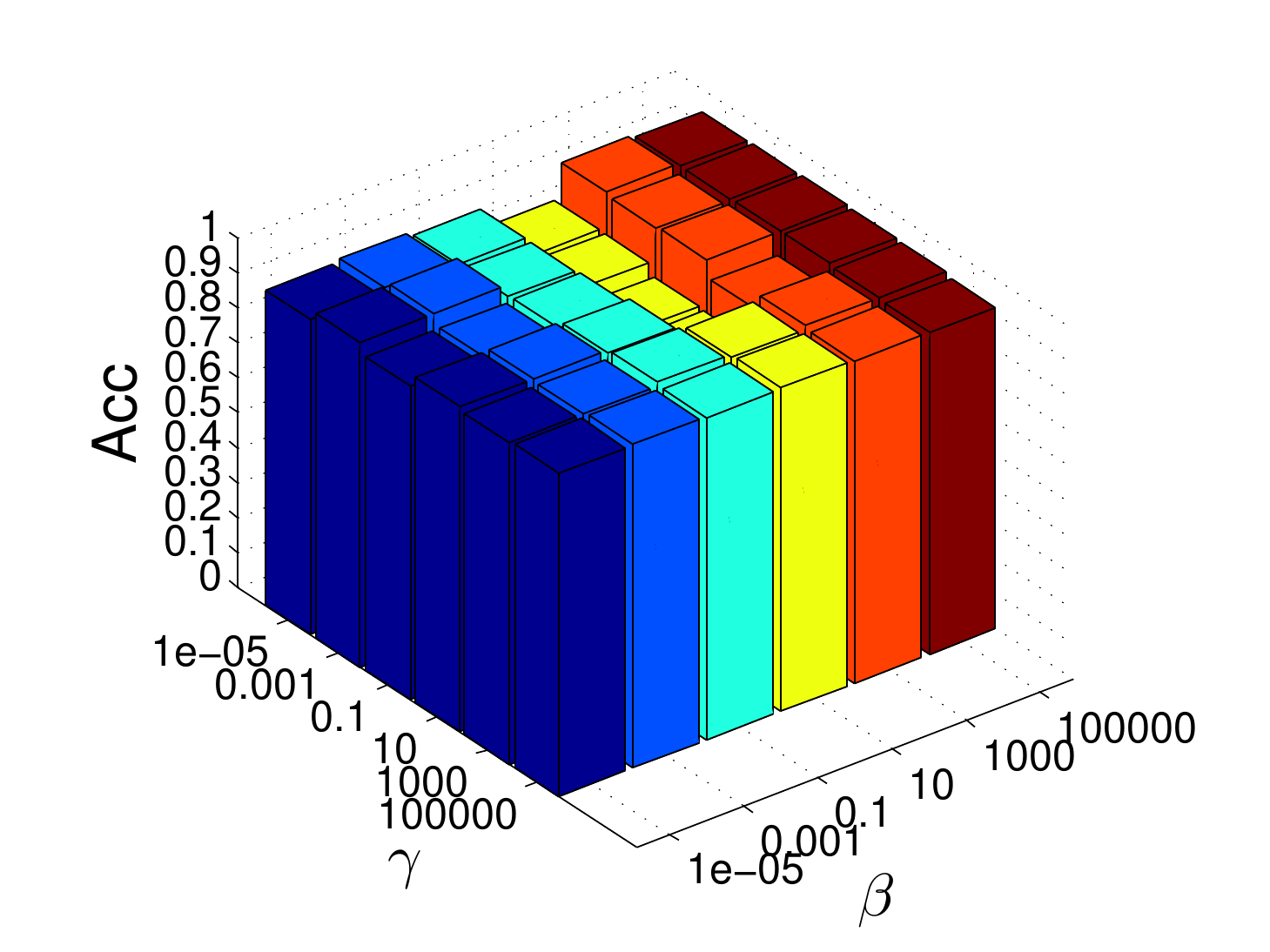}}
\subfloat[$\alpha=10^{-2}$]{\includegraphics[width=.49\textwidth]{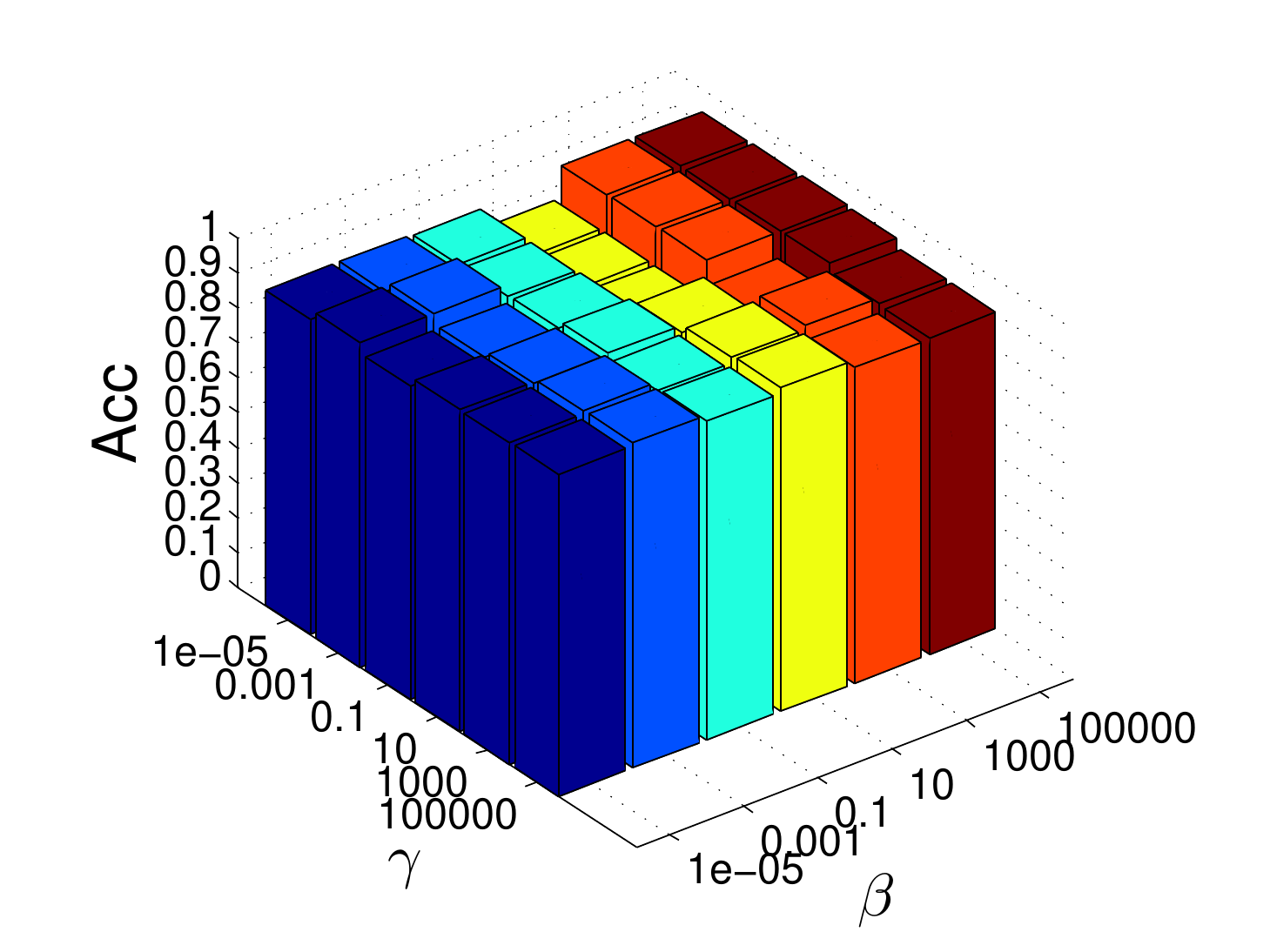}}
\caption{The clustering accuracy of LKGr on JAFFEE Data w.r.t. $\gamma$ and $\beta$.}
\end{figure*}

\begin{figure*}
\centering
\subfloat[$\alpha=10^{-5}$]{\includegraphics[width=.49\textwidth]{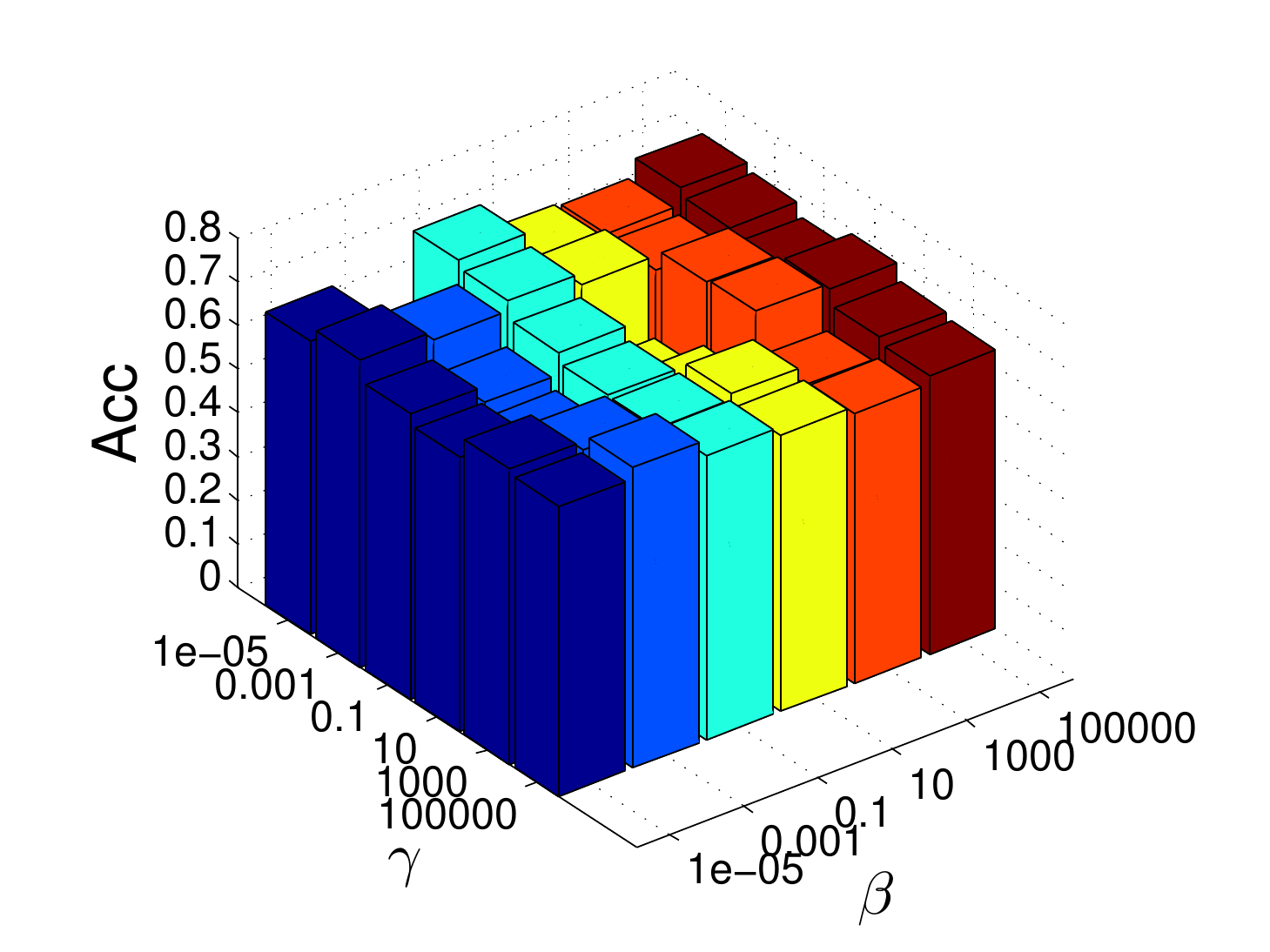}}
\subfloat[$\alpha=10^{-2}$]{\includegraphics[width=.49\textwidth]{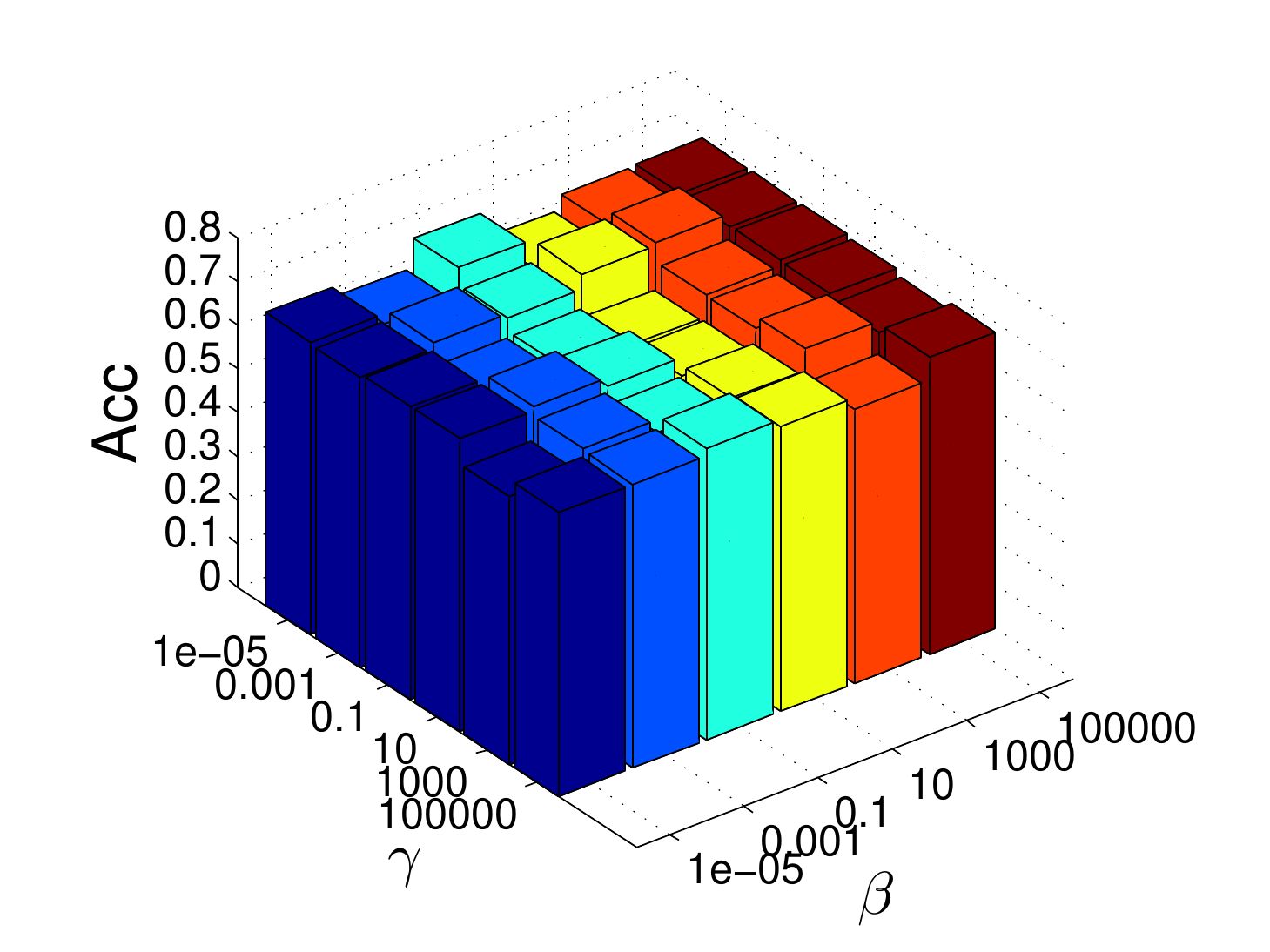}}
\caption{The clustering accuracy of LKGr on ORL Data w.r.t. $\gamma$ and $\beta$.}
\label{orllkgr}
\end{figure*}


For the compared methods, we either use their existing parameter settings or tune them to obtain the best performances. In particular, we can directly obtain the optimal results for KKM, SC, RKKM, MKKM, AASC, and RMKKM methods by implementing the package in \cite{du2015robust}. SSR is a parameter-free model. Hence we only need to tune the parameters for SCSK and SCMK. The experimental results are presented in Table \ref{clusterres}. In most cases, our proposed method LKG achieves the best performance among all state-of-the-art algorithms. In particular, we have the following observations. 
\begin{enumerate}
\item For non-multiple kernel based techniques, we see big differences between the best and average results. This validates the fact that the selection of kernel has a big impact on the final results. Therefore, it is imperative to develop multiple kernel learning method. 
\item As expected, multiple kernel methods work better than single kernel approaches. This is consistent with our belief that multiple kernel methods often exploit complementary information. 

\item Graph-based clustering methods often perform much better than K-means and its extensions. As can be seen, SSR, SCSK, SCMK, and LKG improve clustering performance considerably.
\item By comparing the performance of SCMK and LKG, we can clearly see the advantage of our low-rank kernel learning approach. This demonstrates that it is beneficial to adopt our proposed kernel learning method.
\end{enumerate}

\begin{table}[!htbp]
\centering
\caption{Wilcoxon Signed Rank Test on all Data sets.}
\label{Wilcoxon}
\resizebox{1.\textwidth}{!}{
\begin{tabular}{|c|c|c|c|c|c|c|c|c|c|c|}
\hline
\textrm{Method}&\textrm{Metric}&\textrm{KKM}&\textrm{SC}&\textrm{RKKM}&\textrm{SSR}&\textrm{SCSK}&\textrm{MKKM}&\textrm{AASC}&\textrm{RMKKM}&\textrm{SCMK}\\\hline
\multirow{2}{4em}{\textrm{LKGs}}&Acc&.0039&.0078&.0039&.0117&.3008&.0039&.0039&.0078&.5703\\ 
\cline{2-11}
&NMI&.0039&.0039&.0039&.0078&.2031&.0039&.0039&.0039&.5703\\ \hline
\multirow{2}{4em}{\textrm{LKGr}}&Acc&.0039&.0078&.0039&.0273&.3008&.0039&.0039&.0039&.4268\\
\cline{2-11}
&NMI&.0039&.0039&.0039&.0195&.0547&.0039&.0039&.0078&.3008\\ \hline
\end{tabular}}
\end{table}
To see the significance of improvements, we further apply the Wilcoxon signed rank test to Table \ref{clusterres}. We show the $p$-values in Table \ref{Wilcoxon}. We note that the testing results are under 0.05 when comparing LKGs and LKGr to all other methods except SCSK and SCMK, which were proposed in 2017. Therefore, LKGs and LKGr outperform KKM, SC, RKKM, SSR, MKKM, AASC, and RMKKM with statistical significance.

\subsection{Parameter Sensitivity}
There are three hyper-parameters in our model: $\alpha$, $\beta$, and $\gamma$. To better see the effects of $\beta$ and $\gamma$, we fix $\alpha$ with $10^{-5}$ and $10^{-2}$, and search $\beta$ and $\gamma$ in the range $[10^{-5},\hspace{.1cm} 10^{-3},\hspace{.1cm} 10^{-1},\hspace{.1cm} 10\hspace{.1cm}, 10^3 \hspace{.1cm},10^5]$.  We analyze the sensitivity of our model LKGr to them by using YALE, JAFFE, and ORL data sets as examples, in terms of accuracy. Figures \ref{yalelkgs} to \ref{orllkgr} show our model gives reasonable results in a wide range of parameters. 
\subsection{Examination on Multi-view Data}
Nowadays, data of multiple views are prevailing. Hence, we test our model on multi-view data in this subsection. We employ two widely used multi-view data sets for performance evaluation, namely, Cora \cite{sen2008collective} and NUS-WIDE \cite{chua2009nus}. Note that most of the data sets used in this paper have imbalanced clusters. For example, there are 818, 180, 217, 426, 351, 418, 298 samples in Cora for each cluster, respectively. For clustering, imbalance issue is seldom discussed \cite{zhu2017entropy}. Hence we expect that our method can work well in general circumstances. To do a comprehensive evaluation, more measures, including F-score, Precision, Recall, Adjusted Rand Index (ARI), Entropy, Purity, are used here. Each metric characterizes different properties for the clustering. Except for entropy, the other metrics with a larger value means a better performance.

We implement the algorithms on each view of them and report the clustering results in Table \ref{cora} and \ref{wide}. For our algorithms LKGs and LKGr, we repeat 20 times and show the mean values and standard deviations. As can be seen, our approach performs better than all the other baselines in most measures. It is unsurprising that different views give different performances. Our proposed method can work well in general.
\captionsetup{position=top}
\begin{table*}[!ht]
\centering
\renewcommand{\arraystretch}{1.1}
\setlength{\tabcolsep}{.1pt}
\subfloat[1st view\label{1st}]{
\resizebox{1.0\textwidth}{!}{
\begin{tabular}{ |l  |c |c| c| c |c| c| c| c| c| c| c|c|c }
\hline
Methods 	& F-score	   & Precision    & Recall 	&NMI     & ARI      & Entropy  &  Acc  & Purity   \\	\hline
KKM  	&.306(.302) & .186(.183) & .993(.891)&.108(.070)  & .015(.008)    &.863(.419)   & .341(.322) & .357(.335) \\	\hline
SC	& .304(.289) &.192(.181) &  \textbf{.995}(.772)&.128(.026)  & .028(.004)    & 2.269(.707)  & .344(.295)& .370(.312) \\	\hline
 MKKM & .282 & .194 & .525  &.172& .029    & 1.688   & .349& .402 \\	\hline
AASC	& .293 & .178& .836&.044 &-.004    &  .614 &.290 & .312 \\	\hline
RMKKM&.311& .190 & .859&.141  & .025    & .634   & .361 & .376 \\	\hline
LKGs	& .303(0) & .179(0) & .989(.001)&.005(.001)  & 0(0)    & \textbf{.062}(.003)   & .302(0) & .303(0)\\	\hline
LKGr& \textbf{.335}(.012) & \textbf{ .326}(.015) &.346(.027)& \textbf{.298}(.008)  &  \textbf{.184}(.014)    & 2.536(.103)   &  \textbf{.405}(.016) &  \textbf{.499}(.013) \\	\hline
\end{tabular}}
}\\
\subfloat[2nd view\label{1st}]{
\resizebox{1.0\textwidth}{!}{
\begin{tabular}{ |l  |c |c| c| c |c| c| c| c| c| c| c|c|c }
\hline
Methods 	& F-score	   & Precision    & Recall 	&NMI     & ARI      & Entropy  &  Acc  & Purity   \\	\hline
KKM  	&.304(.268) & .264(.215) & .996(.515)&.169(.080)  & .103(.045)    &2.686(1.614)   & .359(.313) & .416(.350) \\	\hline
SC	& .301(.271) &.183(.180) & .930(.641)&.045(.017)  & .006(.001)    &2.025(1.041)  & .300(.269)& .323(.308) \\	\hline
 MKKM & .246 & .259 & .235  &.147& .091    & 2.707   & .330& .392 \\	\hline
AASC	& .301 & .180& .922&.006 &.002   & .299 &.300 & .305\\	\hline
RMKKM&.264& .271 & .259&.171  & .108    & 2.636   & .358 & .415 \\	\hline
LKGs	& .304(0) & .180(0) & \textbf{.997}(0)&.005(0)  & .001(0)    & \textbf{.028}(0)  & .304(0) & .304(0) \\	\hline
LKGr&\textbf{ .340}(.006) & \textbf{.351}(.010) & .330(.009)&\textbf{.280}(.004)  & \textbf{.201}(.009)    & 2.675(.033)   & \textbf{.452}(.009) &\textbf{ .517}(.006) \\	\hline
\end{tabular}}
}\\

\caption{Performance of various clustering methods on the Cora data set. \label{cora}}
\end{table*}

\begin{table*}[!ht]
\centering
\renewcommand{\arraystretch}{1.1}
\setlength{\tabcolsep}{.1pt}
\subfloat[1st view\label{1st}]{
\resizebox{1.0\textwidth}{!}{
\begin{tabular}{ |l  |c |c| c| c |c| c| c| c| c| c| c|c|c }
\hline
Methods 	& F-score	   & Precision    & Recall 	&NMI     & ARI      & Entropy  &  Acc  & Purity   \\	\hline
KKM  	&.416(.399) & .393(.338) & .939(.524)&.242(.195)  & .202(.133)    &.1.898(1.558)   & .501(.438) & .533(.483) \\	\hline
SC	& .459(.407) &.391(.287) & \textbf{.992}(.796)&.202(.073)  & .212(.059)    & 1.821(.665)  & .529(.368)& .530(.375) \\	\hline
 MKKM & .401& .351 & .475  &.231& .155  & 1.719   & .431& .508 \\	\hline
AASC	& .381 & .263& .692&.101 &.020   & \textbf{.825} &.351& .354 \\	\hline
RMKKM&.408& .378 & .448&.256  & .185    & 1.845   & .450 & .552 \\	\hline
LKGs	&  \textbf{.460}(.036) & .408(.060) & .523(.024)& \textbf{.303}(.580)  & .234(.075)    & 1.743(.192)   & .518(.039) & \textbf{.556}(.050) \\	\hline
LKGr& .449(.003) & \textbf{.421}(.009) & .480(.007)&.224(.009)  & \textbf{.244}(.009)    & 1.881(.032)   &  \textbf{.543}(.014) & .544(.010) \\	\hline
\end{tabular}}
}\\

\subfloat[2nd view\label{2nd}]{
\resizebox{1.0\textwidth}{!}{
\begin{tabular}{ |l  |c |c| c| c |c| c| c| c| c| c| c|c|c }
\hline
Methods 	& F-score	   & Precision    & Recall 	&NMI     & ARI      & Entropy  &  Acc  & Purity   \\	\hline
KKM  	&.428(.397) & .425(.335) & \textbf{.986}(.532)&.269(.203)  &.231(.126)    &1.973(1.533)   & .497(.439) &.547(.477) \\	\hline
SC	& .428(.398) &.359(.270) & .985(.817)&.178(.057)  & .179(.030)    & 1.658(.612)  & .490(.354)& .490(.358) \\	\hline
 MKKM & .404 & .382 & .429 &.248& .184   & 1.885  & .479& .521 \\	\hline
AASC	& .368& .265& .605&.087 &.023& \textbf{1.184}&.340& .363\\	\hline
RMKKM&.439& .416& .465&.287 & .233  & 1.892   & .482& \textbf{ .571} \\	\hline
LKGs	&  \textbf{.478}(.013) & \textbf{ .449}(.020) & .512(.016)& \textbf{.315}(.028)  &  \textbf{.283}(.021)    & 1.850(.067)   &  \textbf{.548}(.014) & .568(.011) \\	\hline
LKGr&.439(.025) & .434(.035) & .445(.014)&.286(.021)  & .244(.040)    & 1.958(.047)   & \textbf{.548}(.030) & .558(.025) \\	\hline
\end{tabular}}
}\\

\caption{Performance of various clustering methods on the NUS-WIDE data set. \label{wide}}
\end{table*}

\section{Conclusion}
\label{conclusion}
In this paper, we propose a multiple kernel learning based graph clustering method. Different from the existing multiple kernel learning methods, our method explicitly assumes that the consensus kernel matrix should be low-rank and lies in the neighborhood of the combined kernel. As a result, the learned graph is more informative and discriminative, especially when the data is subject to noise and outliers. Experimental results on both image clustering and document clustering demonstrate that our method indeed improves clustering performance compared to existing clustering techniques.
\section*{Acknowledgments}
This paper was in part supported by Grants from the Natural
Science Foundation of China (Nos. 61806045, 61572111, and 61772115), a 985 Project
of UESTC (No. A1098531023601041), three Fundamental
Research Fund for the Central Universities of China (Nos.
A03017023701012, ZYGX2017KYQD177, and ZYGX2016J086), and the China Postdoctoral Science Foundation (No. 2016M602677).
\section*{References}
\bibliographystyle{elsarticle-num}
\bibliography{ref}

\end{document}